\documentclass[preprint,authoryear,12pt]{elsarticle}

\usepackage{amssymb,amsmath}
\usepackage{graphicx}
\usepackage[table,xcdraw]{xcolor}

\usepackage{lineno}

\usepackage{makecell}
\usepackage{multirow}

\usepackage[hidelinks]{hyperref}
\usepackage{cite}
\usepackage{xcolor}
\usepackage{color}
\hypersetup{
    colorlinks,
    linkcolor={red!50!black},
    citecolor={blue!50!black},
    urlcolor={blue!80!black}
}

\begin{document}
\begin{frontmatter}

\title{3D convolutional neural network for abdominal aortic aneurysm segmentation}
\author[add1,add2,add3]{Karen~L\'opez-Linares\corref{mycorrespondingauthor}}
\cortext[mycorrespondingauthor]{Equally contributing, corresponding authors}\ead{klopez@vicomtech.org}
\author[add1,add2]{Inmaculada~Garc\'ia\corref{mycorrespondingauthor}}\ead{igarcia@vicomtech.org}
\author[add2,add5]{Ainhoa~Garc\'ia-Familiar}
\author[add1,add2]{Iv\'an~Mac\'ia}
\author[add3,add4]{Miguel~A.~Gonz\'alez~Ballester}

\address[add1]{Vicomtech Foundation, San Sebasti\'an, Spain}
\address[add2]{Biodonostia Health Research Institute, San Sebasti\'an, Spain}
\address[add3]{BCN MedTech, Dept. of Information and Communication Technologies, Universitat Pompeu Fabra, Barcelona, Spain}
\address[add4]{ICREA, Barcelona, Spain}
\address[add5]{Donostia University Hospital, San Sebasti\'an, Spain}

\begin{abstract}
An abdominal aortic aneurysm (AAA) is a focal dilation of the aorta that, if not treated, tends to grow and may rupture. A significant unmet need in the assessment of AAA disease, for the diagnosis, prognosis and follow-up, is the determination of rupture risk, which is currently based on the manual measurement of the aneurysm diameter in a selected Computed Tomography Angiography (CTA) scan. However, there is a lack of standardization determining the degree and rate of disease progression, due to the lack of robust, automated aneurysm segmentation tools that allow quantitatively analyzing the AAA. In this work, we aim at proposing the first 3D convolutional neural network for the segmentation of aneurysms both from preoperative and postoperative CTA scans. We extensively validate its performance in terms of diameter measurements, to test its applicability in the clinical practice, as well as regarding the relative volume difference, and Dice and Jaccard scores. The proposed method yields a mean diameter measurement error of 3.3~mm, a relative volume difference of 8.58\%, and Dice and Jaccard scores of 87\% and 77\%, respectively. At a clinical level, an aneurysm enlargement of 10~mm is considered relevant, thus, our method is suitable to automatically determine the AAA diameter and opens up the opportunity for more complex aneurysm analysis.   
\end{abstract}

\begin{keyword}
    Segmentation \sep Deep learning \sep Abdominal aortic aneurysm \sep Endovascular aneurysm repair
\end{keyword}

\end{frontmatter}


\section{Introduction}
\label{sec:intro}
An Abdominal Aortic Aneurysm (AAA) is the condition consisting in the weakening and ballooning of the abdominal region of the aorta, and it is the primary cause of the significant death and morbidity attributed to arterial aneurysm disease~\citep{Rutherford2018}. An AAA arises as a result of a failure of the major structural proteins of the aorta (elastin and collagen), mostly developing after a degeneration of the media that may lead to widening of the vessel lumen and loss of structural integrity. In the infrarenal abdominal aorta, an aneurysm is  defined when the maximum abdominal aortic diameter is larger than 30~mm or it exceeds the normal diameter in more than 50\%, usually assessed in a Computed Tomography Angiography (CTA) image. Without treatment, the aneurysm tends to grow since it is unable to withstand the forces of the luminal blood pressure, resulting in progressive dilatation, and eventually, rupture. 

Current practice guidelines for the management of an AAA state that the treatment depends on the size or diameter of the aneurysm and the balance between the risk of aneurysm rupture and the operative mortality~\citep{Moll_Powell}. Lately, the treatment of abdominal aortic aneurysms has shifted from open surgery to a minimally invasive alternative known as Endovascular Aneurysm Repair (EVAR). EVAR involves the deployment and fixation of a stent graft using a catheter, introduced through the femoral arteries. This procedure excludes the damaged aneurysm wall from blood circulation, creating a thrombus that shrinks after a successful intervention. Even if it has shown benefits with respect to open surgery, in EVAR the aneurysm is excluded from blood circulation but it is not removed. In the long term, this may lead to complications, and it is therefore required that patients treated with EVAR undergo lifelong surveillance to assess their progress and prognosis postoperatively. This follow-up consists in examining CTA scans taken at least yearly to detect potential complications and to measure the 2D maximum AAA diameter. A reduced aneurysm diameter means a positive progression, whereas an enlarged aneurysm requires further follow-up or reintervention depending on the risk of rupture.   

However, a significant unmet need in the assessment of AAA disease, both preoperatively and postoperatively, is the determination of rupture risk. Maximal AAA diameter is the standard basis for predicting it, with larger diameters associated with increasing risk of rupture. The growth rate must also be considered, but it is challenging to assess on a consistent basis. Guidelines indicate intervention when the aneurysm diameter exceeds 55~mm or when the rate of expansion is greater than or equal to 10~mm in a 12-month period~\citep{Rutherford2018}.  

According to the Society of Vascular Surgery practice guidelines~\citep{Chaikof_Dalman}, there is a lack of standardization determining the degree and rate of disease progression, and a significant variability exists when measuring and reporting the diameter value: the diameter is measured manually on a certain user-selected slice, and significant intra- and inter-observer variability exists. This standardization is hindered by the lack of automatic aneurysm or thrombus segmentation algorithms that allow precise measurement of thrombus maximum diameter. Furthermore, there are other features of AAAs that correlate with increased risk of rupture, such as aneurysm morphology or volume~\citep{Rutherford2018}, which cannot be evaluated if the AAA is not properly segmented. 

Hence, this work aims at proposing a 3D Convolutional Neural Network (CNN) for AAA segmentation from preoperative and postoperative CTA images. It opens up the opportunity for standardizing AAA diameter measurements, as well as providing the basic tool for a more complex analysis of the evolution of AAA geometry. The performance of the 3D segmentation CNN is tested against a 2D approach with a large number of preoperative and postoperative datasets. It yields a mean Dice similarity coefficient of 87\%, and a mean maximum diameter difference of 3.3~mm, which falls below the threshold of 10~mm considered as clinically relevant~\citep{Rutherford2018}.


 
\section{Literature review}
\label{sec:soa}
\begin{figure}
    \centering
    \includegraphics[width=\linewidth]{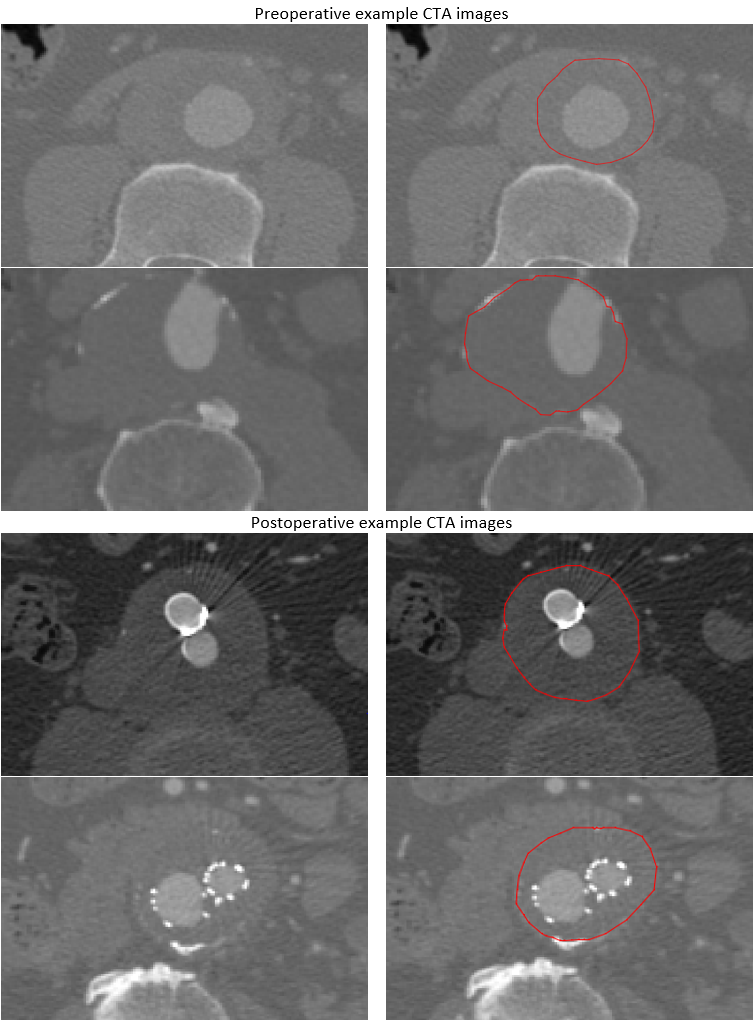}
    \caption{Example preoperative and postoperative CTA images (left) and corresponding outer wall annotations (right) showing the challenges when segmenting the aneurysm: similar intensity to adjacent structures, metal artifacts in the postoperative data, and varying shape and sizes.}
    \label{fig:challenges}
    \end{figure}

Traditionally, aneurysm segmentation has been addressed with intensity-based semi-automatic algorithms (level-sets, active shape models, graph cuts) combined with shape priors. Purely intensity-based techniques fail to correctly detect the non-contrasted outer thrombus boundaries, since there are adjacent structures that have similar intensity values (see Figure~\ref{fig:challenges}) into which the segmentation tends to overflow. With the insertion of a shape constraint this leakage can be further controlled. This is the case of the approximations proposed in~\citep{Mot10,kyu10,Duq12,Egg11, Siriapisith_2018}, where the authors modify 2D or 3D graph-cut energy minimization functions by adding shape models to reduce the overflow into adjacent structures for preoperative aneurysm segmentation. In all of these methods, prior lumen or centerline extraction~\citep{Mot10,Egg11,Siriapisith_2018}, or manual user initialization are required~\citep{kyu10, Duq12}, as well as extensive parameter optimization. A similar approximation but based in the 2D level-set equation is presented in~\citep{Zoh12}, where several criteria are proposed to stop the evolution of the level-set curve to avoid leakage and with a strong assumption on the presence of calcifications, again only for preoperative data. Radial model based approximations that assume an almost circular shape of the thrombus have also been presented in~\citep{Mac09}. In~\citep{dem09} and~\citep{Lalys_Yan_Kaladji} deformable model-based approaches are described, in which an initial contour extracted from the lumen segmentation is deformed according to an energy function until it sticks to the thrombus boundaries. ~\citep{Lalys_Yan_Kaladji} is one of the few studies analyzing both preoperative and postoperative data. Even if some of these methods provide good results, they require the optimization of many parameters and are dataset-dependant, which reduces the robustness and the reproducibility required in a real clinical setting. Most of them also rely on user interaction and/or prior lumen segmentation along with centerline extraction. 

Regarding machine learning and deep learning-based approaches, several works have been proposed. In~\citep{Mai12, Mai14}, aneurysm segmentation is performed with active learning and supervised random forest classifiers. The methods are semi-automatic and no a priori geometric models are needed. In~\citep{Mai12} the segmentation problem is addressed as a slice-by-slice multi-class classification of pixel samples. First, active learning techniques are used to select optimal feature sets, evaluating the information gain of a variety of intensity based features, to train a random forest classifier and perform voxel-based segmentation, which takes about 22 minutes in total.  User interaction is needed during active learning, such that at some iterations previously misclassified data samples are added to the training set and morphological operations are required to refine the segmentation. The work in~\citep{Mai14} is an extension of the previous method, where new features are added for the classification. In both cases, classification accuracy is evaluated, but no comparison with a 3D ground truth segmentation is reported. Both methods apply to postoperative CTA data. Recently, in~\citep{Aik16} a novel and automatic approach to preoperative AAA region detection and segmentation is described, based on deep belief networks. The detection is done in 2D, patch-wise, with patches coming from a unique dataset. Two networks are proposed: one detects large aneurysm patches; the other, detects small aneurysm patches, bones, organs and air. For the segmentation, another deep belief net is trained with 40 image patches containing aneurysm. A comparison with the ground truth is not provided. In~\citep{Zheng2018}, a 2D slice-by-slice U-net network is trained relying on a strong data augmentation, since only 3 datasets are available. As shown in this literature review, most of the state-of-the-art approaches apply only to preoperative or postoperative data separately, require additional inputs in the form of segmentation of other structures or user interaction, and are parametric approaches. 

In previous works, we proposed the use of an adapted Hollistically-Nested Edge Detection (HED) network trained in 2D, slice-by-slice, to segment the AAA. In~\citep{Lopez}, the aneurysm region of interest was firstly detected from a whole postoperative CTA volume using a neural network. This region was then passed to the HED network in order to run the segmentation. The pipeline was fully automated and was tested using 13 postoperative CTA datasets. In~\citep{klopez_miccai}, we trained the same network also for preoperative data, for a total of 38 datasets. The advantage of this method is that it is non-parametric and does not require any prior lumen segmentation or centerline extraction. Hence, our current goal is to extend the adapted HED segmentation network to 3D, designing the first 3D convolutional neural network (CNN) for aneurysm segmentation. Our approach is also one of the few working for preoperative and postoperative data. Additionally, compared to the previous 2D approach, we optimize the processing pipeline and we propose a more appropriate loss function for the 3D setup, which is based on a weighted Dice coefficient. Finally, we aim at extensively validating its performance for a large number of preoperative and postoperative scans from a clinical point of view.  

\section{Methodology}
\label{sec:methods}

This section describes the methodology employed for preoperative and postoperative 3D AAA segmentation. The employed datasets are described in Section~\ref{sec:mat}, as well as how ground truth annotations have been generated. Section~\ref{sec:seg} presents the segmentation pipeline, which is based on the use of a 3D Convolutional Neural Network (CNN), extended from our previous work presented in~\citep{Lopez}. In order to compare the results of the 3D segmentation with a 2D approach, the proposed network is trained both in 2D and 3D with the same training setup. Finally, the same networks are trained also separately for preoperative and postoperative data in order to select the best approach. 

\subsection{Materials}
\label{sec:mat}
The imaging data for this study have been provided by Donostia University Hospital and consist of 78 preoperative and postoperative contrast-enhanced CTA datasets from 54 different patients presenting an infrarenal aneurysm. The datasets have been obtained with scanners from different manufacturers and models: Toshiba Aquilion, GE Lightspeed RT16, GE optima 264 CT660 and GE Lightspeed VCT. They have varying spatial resolution, ranging from 0.72-0.97 in x and y direction, and 0.625-0.8 in z direction.

CTA volumes have been divided into two groups: 50 scans from 34 patients for training and validating our algorithm, and 28 scans from 20 patients for testing it. From the training dataset, 22 volumes correspond to preoperative scans, while 28  are postoperative images. Similarly, for the testing dataset 12 scans are from the preoperative stage, whereas 16 are postoperative volumes. If for the same patient the preoperative and postoperative scans are available, both images are included in the same set. For all the images, ground truth annotations have been generated semi-automatically by trained experts, using a previously developed segmentation method~\citep{Lopez} and editing the results. Table~\ref{tab:data} summarizes the employed datasets.

\begin{table}[]
\centering
\begin{tabular}{llll}
                                                                     & Total      & Pre     & Post     \\
\multirow{2}{*}{\begin{tabular}[c]{@{}l@{}}Train\\ CTA\end{tabular}} & 50         & 22      & 28       \\
                                                                     & \multicolumn{3}{c}{34 patients} \\
\multirow{2}{*}{\begin{tabular}[c]{@{}l@{}}Test\\ CTA\end{tabular}}  & 28         & 12      & 16       \\
                                                                     & \multicolumn{3}{c}{20 patients}
\end{tabular}
\caption{Summary of the employed data }
    \label{tab:data}
\end{table}

\subsection{Segmentation pipeline}
\label{sec:seg}
\begin{figure}
    \centering
    \includegraphics[width=\linewidth]{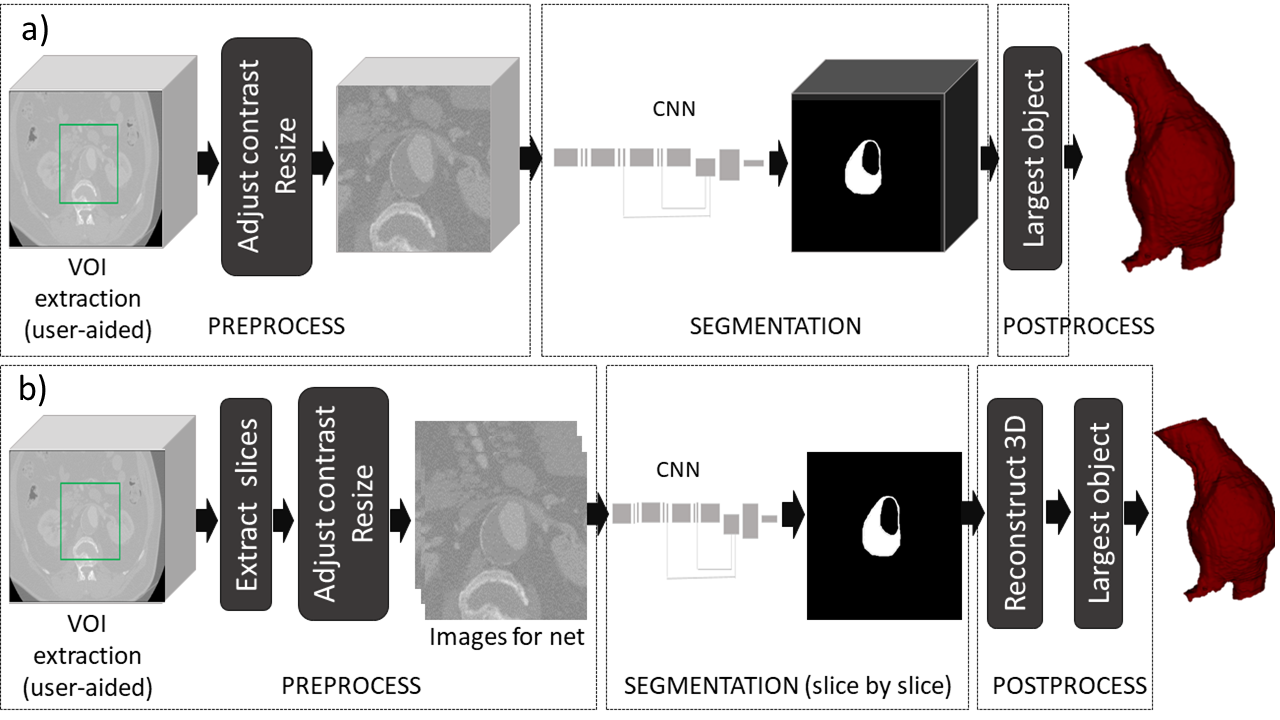}
    \caption{Pipeline for segmentation using 3D CNN (a) and a 2D CNN (b).}
    \label{pipelines}
    \end{figure}

 The following pipeline is applied to segment the AAA from CTA images: (1) image pre-processing, which includes extraction of the volume of interest, contrast enhancement, and resizing; (2) segmentation of the thrombus using the CNN; (3) image post-processing to binarize the network's prediction and to refine the segmentation results by removing small objects. In order to compare the results obtained using the 3D approach with a 2D method, the same setup is applied in 2D, but with two additional steps. These steps refer to the extraction of the 2D slices from the input 3D volume and the reconstruction of the 2D output slices to generate the 3D output segmentation volume. Figure~\ref{pipelines} depicts both pipelines.  
 
\subsubsection{Data preprocessing}

The first step in our pipeline is the extraction of a volume of interest. As abdominal CTA scans span a large body area, the images are manually cropped to contain only a region around the aneurysm in order to limit the amount of memory required for training. Afterwards, image contrast is enhanced by applying a simple window-level adjustment, rescaling the intensity between 0 and 255. For training the 2D CNN, axial slices are extracted from the 3D volume. Finally, 2D slices are resized to 256x256, whereas in 3D the resolution is reduced to 128x128x64 instead of having an axial resolution of 256x256 to deal with the expensive memory requirements of training a 3D CNN.

\subsubsection{Convolutional neural network}
The chosen network architecture is a modification of the Holistically-Nested Edge Detection (HED) network~\citep{He_Girshick}, which was already presented for 2D postoperative CTA segmentation in~\citep{Lopez}. In the current work, this network architecture is extended to perform in 3D.

\begin{figure}
    \centering
    \includegraphics[width=\linewidth]{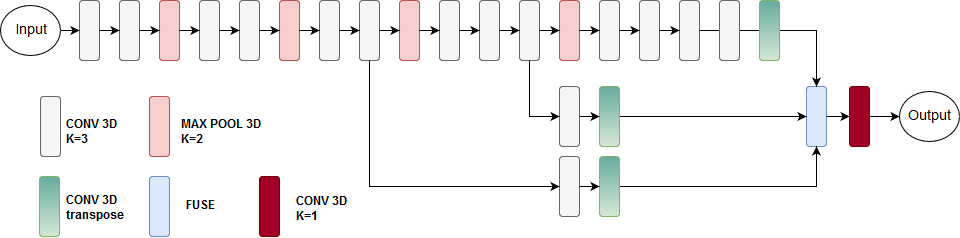}
    \caption{Proposed 3D CNN, extended from~\citep{Lopez} and based on the 2D Hollistically-nested edge detection network~\citep{Xie_Tu_2015}.  }
    \label{fig:net}
\end{figure}

\begin{figure}
    \centering
    \includegraphics[width=\linewidth]{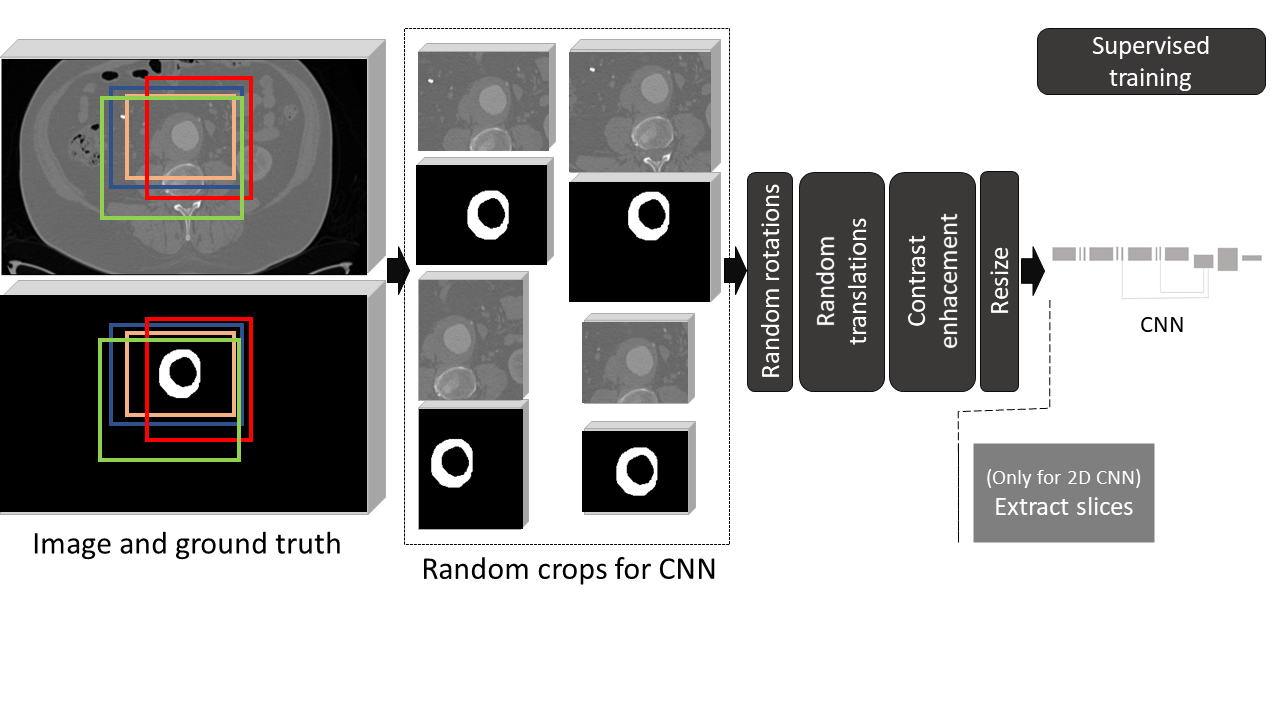}
    \caption{Training scheme. The same training strategy is used for training the 2D and the 3D CNNs, with the only difference being the slice extraction required for training in 2D. }
    \label{fig:training_scheme}
    \end{figure}

\begin{description}
\item[Net architecture:] The proposed network architecture combines edge detection with the preservation of the shape and appearance information of the aneurysm. In traditional solely edge-based and intensity-based segmentation approaches, the inclusion of a shape and appearance prior is essential to improve the aneurysm segmentation problem, since edges between the aneurysm and adjacent structures are commonly diffuse. This knowledge was exploited to propose the 2D architecture in~\citep{Lopez} that combines different relevant scales: low scale for fine edge and border detection; and more global appearance and localization information, obtained with a macro scale, that ensures the smooth contour of the aneurysm. The feature maps in the original HED architecture are extracted from the very beginning of the network and from the output of the last convolutional layer. Shallower connections provide very rich edge information while the final maps are more coarse and fuzzy. However, these first richer maps fail to detect the aneurysm borders, while responding strongly to more contrasted structures, such as the lumen or the vertebrae. On the other hand, the deepest map seems to correctly locate the thrombus area and shape appearance. Hence, as compared to the original HED we removed the first two output side connections and kept only the last ones, that connect the deepest layers. This way, we dealt with one of the main challenges of thrombus segmentation, namely the lack of an apparent edge between the aneurysm and adjacent structures with similar intensities. Furthermore, we removed the initial padding to improve resolution and used element wise fusing instead of concatenation to keep strongest activations. The final architecture, which is now extended to 3D by replacing the 2D operations by its 3D counterparts, is smaller and has less parameters than other architectures in the literature, thus, having lower memory consumption and being more adequate for training with few images. This 3D version of the CNN allows for processing the whole volume at once, thus leveraging the complete three-dimensional context information to have a more coherent and consistent segmentation, as opposed to a 2D architecture in which the image has to be processed slice by slice. Figure~\ref{fig:net} shows our proposed 3D architecture, with details about the layers. 

\item[Loss function:] Given the strong class imbalance between background pixels and foreground (aneurysm) pixels, the use of a weighted Dice loss function is proposed. We compute two Dice coefficients, one taking only into account foreground pixels, the other one considering only the background. Then, a weight of 0.9 is assigned to the foreground Dice score and a weight of 0.1 to the background. Therefore, the function to optimize is:

\begin{equation*}
L_{dice}= 1 - 0.1D_{background}+0.9D_{foreground}
\end{equation*}
Where $D_{background}$ and $D_{foreground}$ are the Dice coefficients computed only for the background and for the foreground, respectively, given the following formula:
\begin{equation*}
  Dice = \frac{2\sum y_{true}\cap y_{predicted}}{\sum y_{true} + \sum y_{predicted} + 1 }  
\end{equation*}

\item[Training approach:] As we have a limited number of training volumes (50 CTA scans), and to make the CNN more robust to variations, data augmentation is applied. Four random crops (which fully contain the aneurysm) are generated from each of the input volumes, to make the network robust against the variability when defining the input volume of interest. Random rotations and translations are then applied to all the cropped volumes, generating 140 volumes per each input scan. With these data, the 2D and 3D version of our network are trained from scratch. We chose this training strategy given the lack of 3D pre-trained models or similar data that could be used for transfer learning. Training from scratch has been proven to be as effective as using a pretrained model, though it can require more time to converge~\citep{He_Girshick}. The networks are trained in a fully supervised manner, providing ground truth pixel-level annotations for each training image. Figure~\ref{fig:training_scheme} summarizes the training process. 

The initial learning rates for both network is set to 1e-4, which is reduced during training by a factor of 0.2 when the validation loss stops improving. For the 2D version, a batch size of 16 images is employed, while for the 3D CNN a batch size of 2 is used due to memory limitations. The Adam optimizer is employed in both cases. Both networks are trained on a TITAN X Pascal (NVIDIA) GPU card with 11.91GB memory.

\end{description}

\subsection{Postprocessing of the results}
The last step consists in processing the network-provided probability maps to obtain the final binary segmentation. Network outputs are binarized using Otsu's thresholding as proposed in~\citep{klopez_isbi}, and only the biggest object is kept in the final mask. 

\section{Results and validation}
 Hereby, we present the results of the 3D aneurysm segmentation pipeline. The performance of the trained model is tested by obtaining the aneurysm segmentation for new 28 CTA volumes, being 12 of them preoperative scans, and 16 of them postoperative. First, the automatically obtained segmentations are compared against ground truth annotations in terms of Dice and Jaccard coefficients. Furthermore, given that the goal is to propose a segmentation approach that can be useful in a clinical setting for the diagnosis and follow-up of patients, relevant clinical measures from the segmentation are also evaluated. These include the maximum axial diameter of the aneurysm and the volume. The manually measured maximum axial diameter is nowadays the standard reference in the clinical practice, so the absolute difference between this diameter, computed automatically from the ground truth, and the automatic segmentation is evaluated. Its precise measurement is essential as it would directly impact the treatment decision. 
 Additionally, the relative volume difference between the ground truth and our segmentation is computed, since the volume is thought to be a better indicator of disease progression than the axial diameter. The results obtained for our new 3D segmentation approach, are finally compared to those obtained with the 2D CNN version. Finally, the same networks are trained and tested separately for the preoperative and the postoperative stages, in order to select the best training approach.  

\subsection{Segmentation accuracy}
First, the following traditional metrics to evaluate the segmentation accuracy have been computed:

\begin{itemize}
    \item Dice coefficient: 
    \begin{equation}
        d = 1 - \frac{2|Y_{true}\cap Y_{pred}|}{|X_{true}|+|X_{pred}|}
    \end{equation}
    \item Jaccard index:
    \begin{equation}
        j = \frac{|Y_{true}\cap Y_{pred}|}{Y_{true}\cup Y_{pred}}
    \end{equation}
\end{itemize}

\begin{figure}
    \centering
    \includegraphics[width=0.75\linewidth]{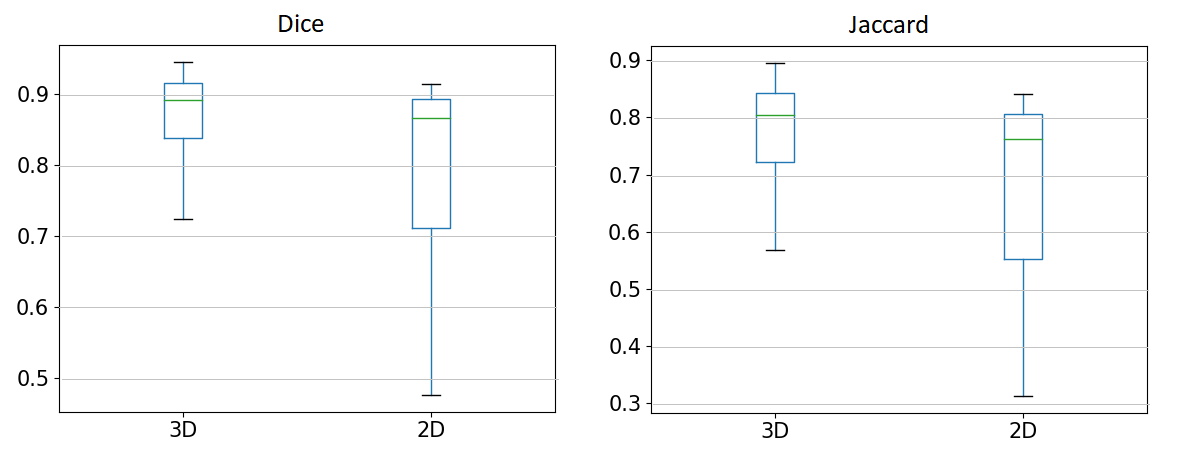}
    \caption{Boxplots for the Dice and Jaccard index obtained for the 2D CNN and 3D CNN.}
    \label{fig:boxplots}
    \end{figure}

Table~\ref{table:summary} summarizes the obtained mean and standard deviation values for both the Dice and the Jaccard scores for the 3D and the 2D networks, trained only with preoperative data, only with postoperative, or with all the scans together. For the novel 3D CNN segmentation approach, we reach a mean Dice coefficient of 87\% when training with all the data, with a 84\% Dice score and a 89\% Dice score values for the preoperative and postoperative data, respectively. The best results are obtained when training with all the data, instead of when training separately for the preoperative and the postoperative. As compared to the 2D approach, a significant improvement (Mann–Whitney U test, p value$<$0.01, one-sided) is observed, being the mean 2D Dice score 80\% when training with all the data. Figure~\ref{fig:boxplots} shows the boxplots for Dice and Jaccard coeficients for the results of the 2D CNN and the 3D CNN.


\begin{table}[]
\begin{tabular}{clll|ll}
              &                 & \multicolumn{2}{c}{3D} & \multicolumn{2}{c}{2D} \\
\thead{Training\\ data} & \thead{Evaluation\\ data} & Dice                   & Jaccard               & Dice                   & Jaccard               \\  \hline
 & \thead{Pre \\Post}    & 0.87$\pm$0.062             & 0.77$\pm$0.093            & 0.80$\pm$0.11              & 0.68$\pm$0.14             \\
              & \thead{Pre}        & 0.84$\pm$0.068             & 0.73$\pm$0.100            & 0.75$\pm$0.14              & 0.62$\pm$0.18             \\
 \multirow{-3}{*}{\thead{Pre\\Post}}              & \thead{Post}       & 0.89$\pm$0.044             & 0.81$\pm$0.069            & 0.84$\pm$0.08              & 0.73$\pm$0.11            \\ \hline
\thead{Pre }     & \thead{Pre}        & 0.81$\pm$0.069             & 0.69$\pm$0.099            & 0.74$\pm$0.12              & 0.60$\pm$0.14             \\ \hline
\thead{Post}     & \thead{Post}       & 0.88$\pm$0.052             & 0.78$\pm$0.077            & 0.80$\pm$0.11              & 0.68$\pm$0.14            
\end{tabular}
\caption{Summary of the quantitative results obtained for aneurysm segmentation using the 3D CNN and the 2D CNN in terms of Dice coefficient and Jaccard index. An evaluation of the differences when training the networks only with preoperative data, only with postoperative or with both altogether is included. }
\label{table:summary}
\end{table}

\subsection{Diameter and volume evaluation}
\begin{figure}
    \centering
    \includegraphics[width=\linewidth]{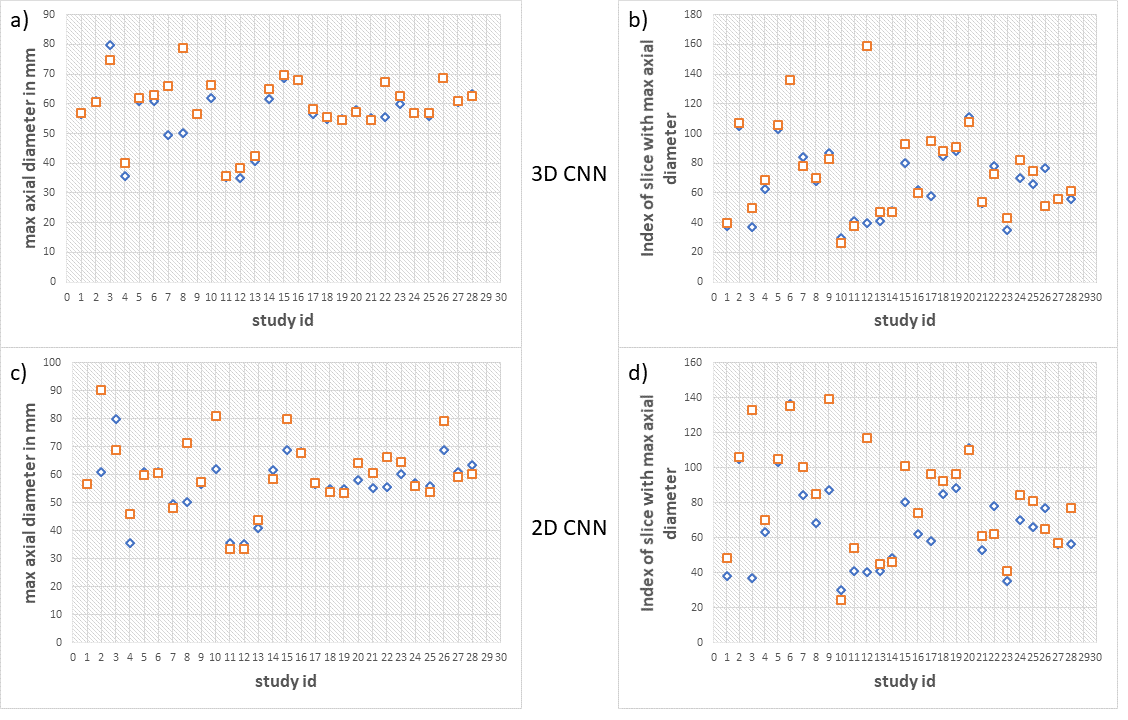}
    \caption{Results of the segmentation in the test set evaluated in terms of maximum axial aneurysm diameter. Ground truth values (obtained from manual segmentation) are shown in blue-rhombus, results from the CNN are in orange-square.  Top: results using the 3D CNN in terms of (a) maximum axial diameter and (b) index of the slice with the maximum diameter. Bottom: segmentation using the 2D CNN in terms of (a) maximum axial diameter and (b) index of the slice with the maximum diameter.  }
    \label{fig:diameters}
    \end{figure}

The maximum axial diameter of the aneurysm is the most important parameter currently employed in the clinical practice for the diagnosis, progression assessment and treatment decision making. 
Therefore, we aim to assess the quality of the segmentation in this regard by comparing the maximum diameter extracted from the ground truth and the automatic segmentation obtained with the 3D and the 2D networks trained with preoperative and postoperative data. For that purpose, we compute the minimum enclosing circle of the aneurysm masks slice by slice, and we select the maximum obtained value. Then, two aspects are evaluated: (1) maximum axial diameter value (mm), and (2), the accuracy finding the slice with the maximum diameter. These results are summarized in Figure~\ref{fig:diameters}.

\begin{figure}
    \centering
    \includegraphics[width=\linewidth]{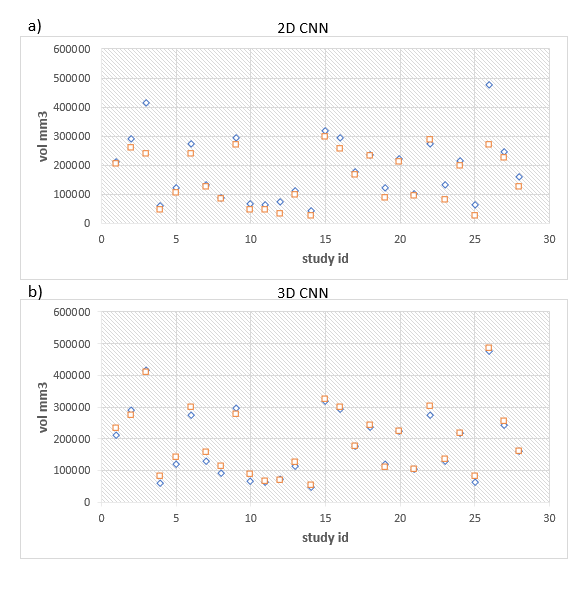}
    \caption{Results of the segmentation in the test set evaluated in terms of aneurysm volume. Ground truth values (obtained from manual segmentation) are shown in blue - rhombus, results from the CNN are in orange - square.  Top: segmentation using 2D CNN (a) Bottom: segmentation using 3D CNN (b)  }
    \label{fig:volume}
    \end{figure}

On the other hand, even if it is not employed in the current clinical practice due to the lack of robust aneurysm segmentation algorithms, the evaluation of the variation in the volume of the aneurysm (instead of in just the diameter in a certain slice) could provide a more suitable biomarker for patient evaluation. Figure~\ref{fig:volume} shows a comparison of the volumes measured from the ground truth aneurysm masks and the segmentations obtained with the 2D and the 3D CNN. The mean relative volume difference is calculated according to the following formula: 
\begin{equation}
    V_{diff} = \frac{1}{i}*\sum_{n=1}^{i}|\frac{V_{pred} - V_{true}}{V_{true}}| , \textit{ where i = 28 test datasets}
    \label{equ:vol_diff}
\end{equation}

Table~\ref{table:rel_vol} presents the results of the relative volume difference and the mean absolute maximum axial diameter difference, for the 3D and the 2D CNN networks trained with preoperative and postoperative data. Again, the 3D approach provides better results for all the metrics than the 2D method.

\begin{table}[]
\begin{tabular}{lcc}
                               & 3D & 2D \\
Mean relative volume difference     & 0.093$\pm$0.09                 & 0.188$\pm$0.161                \\
Mean absolute max diameter difference (mm)  & 3.309$\pm$6.03 & 5.843$\pm$7.098 \\
\end{tabular}
\caption{Mean relative volume difference and mean error measuring the maximum axial diameter between the ground truth and the automatic segmentations generated with the 3D and the 2D CNN. }
\label{table:rel_vol}
\end{table}

\section{Discussion}
\label{sec:dis}

A significant unmet need in the assessment of AAA disease is the evaluation of the aneurysm rupture risk. Maximal AAA diameter is the standard basis for predicting it, but it is challenging to assess its value and evolution without a precise aneurysm segmentation. In this work, we validate the use of a 3D convolutional neural network (CNN) specifically adapted for the segmentation of AAAs from preoperative and postoperative CTA scans. Having this segmentation tool opens up the opportunity for a more complex analysis of the AAA that could help predicting the rupture risk more precisely.

\begin{figure}
    \centering
    \includegraphics[width=\linewidth]{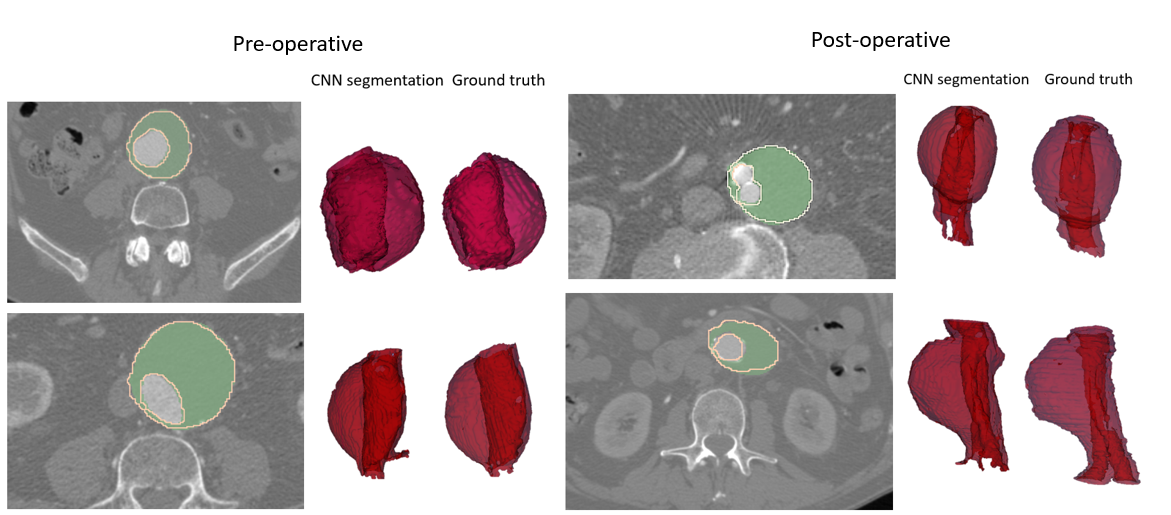}
    \caption{Example of the segmentation results in pre-operative (right) and post-operative (left) CTA volumes. Ground truth (green) and segmentation (orange outline) overlayed on the image (axial slice) and 3D reconstruction.   }
    \label{fig:results_pre_post}
    \end{figure}

In a first experiment, we have trained our 3D CNN with preoperative and postoperative data altogether, and we have compared its testing performance against a network trained only with postoperative or only with preoperative data, in terms of Dice and Jaccard scores. As shown in Table~\ref{table:summary}, training with all the data together provides better results, both for the preoperative and the postoperative data. The difference is even larger for the preoperative AAA segmentation, probably due to a smaller amount of available preoperative scans. Training with all the datasets at once provides the advantage of having more data, which is crucial for deep learning approaches. The most challenging part of AAA segmentation corresponds to correctly locating the outer aortic wall without leaking to adjacent structures, and this can be equally learned from preoperative and postoperative data. The main difference between preoperative and postoperative CTA scans refers to the shape of the aortic lumen: in the preoperative, a unique lumen is observed, whereas in the postoperative scan two lumen regions are visible because of the presence of the stent graft modules. However, since the lumen is always contrasted, learning to find the inner aortic wall corresponding to the edge between the aneurysm and the contrasted lumen is more straightforward. Additionally, our approach is able to overcome the appearance of metal artifacts creating shadows and lights in the thrombosed aneurysm area, providing good segmentation results even under these circumstances.   Figure~\ref{fig:results_pre_post} shows some example results of the 3D CNN tested on preoperative and postoperative data when trained with images from both stages together. 

Secondly, this same experiment has been run by training the 2D version of the CNN slice-by-slice, for comparison purposes. As expected, the results worsen notably when the 3D dimensionality of the image is not taken into account, since some context information is lost and the 3D consistency is not preserved. 

\begin{figure}
    \centering
    \includegraphics[width=\linewidth]{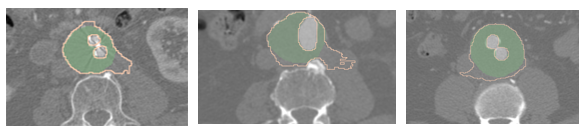}
    \caption{Ground truth (green) and segmentation (orange outline) of the diameter outliers. The first two slices come from the preoperative and postoperative scans of the same patient, whereas the last image corresponds to a different patient. Structures close to the aneurysm have been incorrectly segmented by the 3D CNN.}
    \label{fig:diameter_outliers}
\end{figure}

\begin{figure}
    \centering
    \includegraphics[width=\linewidth]{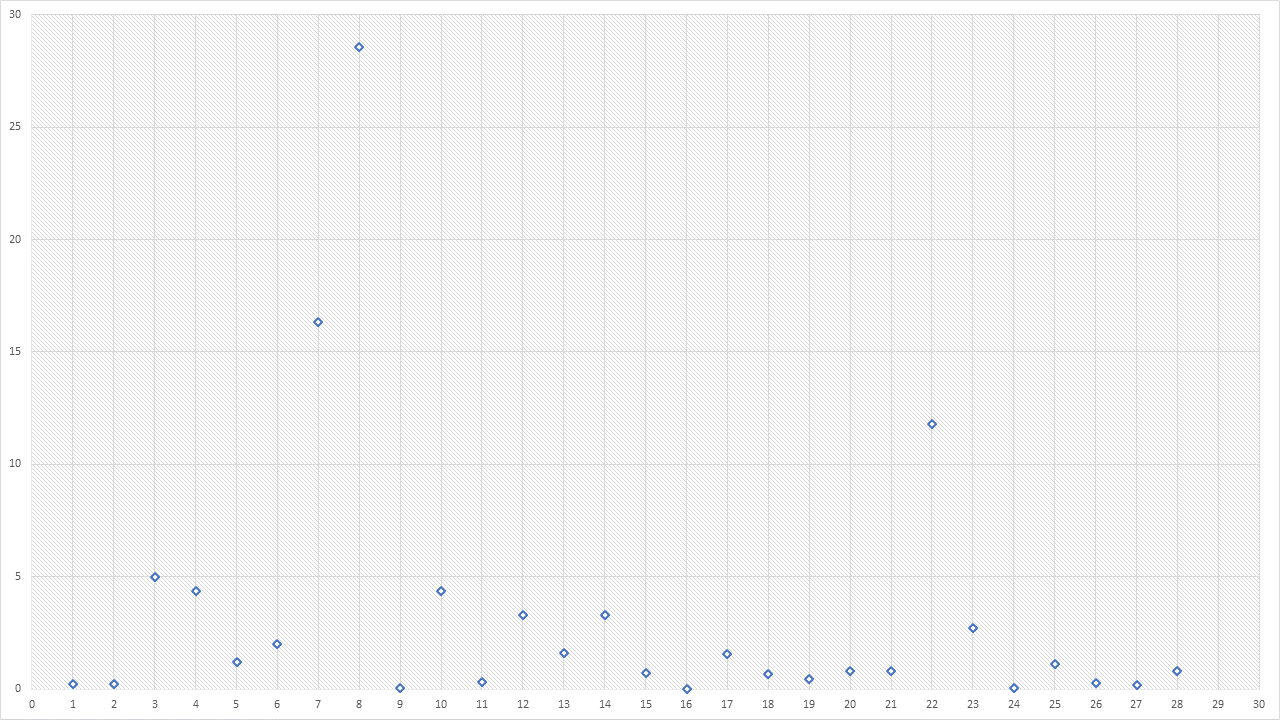}
    \caption{Absolute error, in millimeters, when measuring the maximum axial diameter extracted from the 3D automatic segmentation, as compared to the ground truth. Apart from 3 clear outliers, the error falls below the 10~mm clinical threshold for all the cases.}
    \label{fig:diam_abs_diff}
\end{figure}

Once the best training approach has been selected, which refers to training with preoperative and postoperative data altogether, our proposed 3D approach is validated from a clinical perspective, and compared to the 2D method. For that purpose, we first compute the error evaluating the maximum axial diameter, which is the current standard measurement in the clinical practice. Our 3D method yields a mean error measuring the maximum axial diameter of only 3.309~mm, with a standard deviation of 6.03~mm. Guidelines for aneurysm management state that intervention is required when the aneurysm diameter exceeds 55~mm or when the rate of expansion is greater than 10~mm in a 12-month period~\citep{Chaikof_Dalman}. A diameter enlargement of 10~mm is considered relevant and requires further evaluation or even a re-intervention~\citep{Chaikof_Dalman, Moll_Powell}. Thus, the mean error using our 3D approach falls below the threshold of 10~mm: for 25 out of 28 test cases the error falls below 5~mm. However, three clear outliers can be observed in Figure~\ref{fig:diameters} and more clearly in Figure~\ref{fig:diam_abs_diff}, which make the standard deviation of the error to increase. In these datasets, which correspond to a preoperative and postoperative image of the same patient, and a postoperative image of another different patient, the network provided segmentations leak into adjacent structures with similar intensity, causing the computed diameter values to be larger than the ground truth. This can be observed in  Figure~\ref{fig:diameter_outliers}. Additionally, we evaluate the index of the slice in which the maximum diameter is found, as depicted in Figure~\ref{fig:diameters}. The mean deviation in the index is 10.6 slices, which taking into account the range of the z resolution of the images varies between 6.625-8.48~mm. In a previous experiment presented in~\citep{Lopez}, an inter-observer variability between 1 and 24 slices was found when selecting certain CTA slices, so our current error can be considered acceptable. 

\begin{figure}
    \centering
    \includegraphics[width=0.6\linewidth]{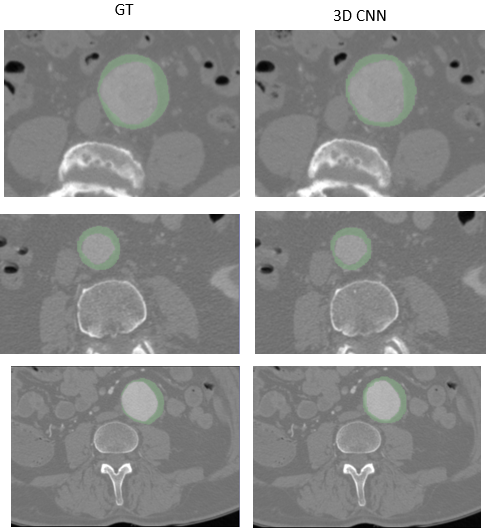}
    \caption{Ground truth (left) and automatic segmentation (right) result for cases with a very small thrombus.}
    \label{fig:small_aneurysm}
\end{figure}

 Finally, we have computed the mean absolute relative volume difference between the ground truth and the automatic 3D segmentation, which could provide a better reference to evaluate aneurysm progression. The obtained mean absolute relative volume difference, computed with Equation~\ref{equ:vol_diff}, is 9.3\%, where the error for preoperative data is 11.2\% and for the postoperative data it is 7.9\%. If small and large aneurysms are evaluated separately, setting the threshold on a volume of 100000~mm3, an error of only 6\% is achieved for large aneurysm, whereas a larger volume difference is attained for smaller ones, around 20\%. This is due to small segmentation error causing larger deviations when the thrombus is small. In some of our cases, the AAA does not practically have a thrombus, and thus the inner and outer aneurysm walls are very close to each other. This can be observed in Figure~\ref{fig:small_aneurysm}.
 Table~\ref{table:compare} shows a comparison between our work and recent approaches in the literature, in terms of the employed data, the mean Dice similarity coefficient and the volume difference, as well as the required inputs for the algorithms. In order to compare our approach to these recent works in the literature, we compute the volume difference as proposed in~\citep{Siriapisith_2018}, which yields a value of 8.58\%, smaller than the one reported in~\citep{Siriapisith_2018}. If the volume difference is only evaluated for preoperative data, our method yields a volume difference of 11.22\%, which is above the value of 9.25\% obtained in~\citep{Siriapisith_2018}. Our approach is only tested against 12 preoperative CTA scans coming from different manufacturers, whereas their approach is defined and optimized with 20 images coming from a single model, and it is specifically designed for the preoperative stage, which can be the reason why they achieved better Dice and volume difference values for the preoperative stage. 
 
 Compared to previous traditional approaches such as~\citep{Siriapisith_2018} and~\citep{Lalys_Yan_Kaladji}, our method has the advantage that it does not rely on any previous segmentation or centerline extraction, which can lead to errors if any of them is not correctly computed. For example, in~\citep{Siriapisith_2018}, the lumen contour needs to be dilated to create the initial contour for the graph-cuts, and the amount of dilation changes between cases, which reduces the robustness and automation of the algorithm. Their algorithm also requires the initial contour to be shifted to avoid falling into the inferior vena cava; separating this structure from the aneurysm is indeed a challenge, since they have similar intensity values and are sometimes touching each other. Additionally, our method has the advantage of being non-parametric as compared to~\citep{Siriapisith_2018} and~\citep{Lalys_Yan_Kaladji}. Regarding the approach in~\citep{Zheng_Zhou_Li_Riga_Yang_2018}, which is also based on a CNN, our approach yields better Dice values, is tested with more cases and works both for the preoperative and the postoperative stage.

\begin{table}[]
\centering
\begin{tabular}{c|cccccc}
Method & \thead{2D\\3D} & \thead{Pre\\Post} & Datasets & Dice  & Vol. diff. & Inputs \\
\hline
\thead{Graph-cuts \\ ~\citep{Siriapisith_2018}} & 3D & Pre & 20 & 91.88\% & 9.25\% & Lumen  \\
\thead{Deformable model\\~\citep{Lalys_Yan_Kaladji}} & 3D & \thead{Pre\\Post} & 145 & 85\% & Not reported & \thead{Lumen\\centerline}  \\
\thead{U-net\\~\citep{Zheng2018}} & 2D & Pre & 3 & 82\% & Not reported & - \\
Current work & 3D & \thead{Pre\\Post} & 80 & 87\% & 8.58\% & - \\

\end{tabular}
\label{table:compare}
\caption{Comparison between our current 3D CNN approach and recent works in the literature.}
\end{table}


\section{Conclusions and future work}
\label{sec:conc}

Hereby, we have proposed a preoperative and postoperative aneurysm segmentation approach based on a 3D convolutional neural network. This network is the 3D extension of a previously designed 2D adaptation of a hollistically-nested edge detection network~\citep{Lopez}, which relies on the combination of feature maps at different scale levels. Low scale features are employed for fine edge and border detection, and more global appearance and localization information is obtained with a macro scale that ensures the smooth contour of the aneurysm, which is located above the vertebrae, around the lumen and between both kidneys. This 3D network, trained to minimize a novel weighted Dice loss function and tested with a large number of preoperative and postoperative CTA scans, has shown to provide better results than its 2D counterpart. 

A validation study to test its usability in the clinical practice is included, which is based on measuring the aneurysm's maximum axial diameter and volume. Our approach yields a mean error of 3.309~mm when measuring the axial diameter, which falls below the threshold of 10~mm change reported as relevant in clinical guidelines~\citep{Rutherford2018,Chaikof_Dalman}. For 25 out of 28 test datasets, the error falls below 5~mm, but 3 outliers make the standard deviation of the error to increase. These guidelines state that a significant unmet need in the assessment of AAA disease, both preoperatively and postoperatively, is the determination of rupture risk, being the maximal AAA diameter the standard basis for predicting it. This method sets the basis for an automatized, objective approximation to its evaluation. According to the mean volume difference, which is considered a better predictor of rupture risk, but it is still not employed in the clinical routine~\citep{Chaikof_Dalman}, an error of 9.3\% is achieved, which is lower than the error reported by other approaches in the literature. It has also the advantage of being non-parametric and versatile, it is robust and works with CTA scans coming from different manufacturers, and does not require prior segmentation of any other structures. These characteristics make it suitable and easily translatable to the clinical practice.            

Regarding future work, we first aim at improving the segmentation of small aneurysms. As explained in Section~\ref{sec:dis}, for aneurysms in which the thrombosed area is very thin our method yields diameter and volume difference errors that should be improved. Besides, multiclass segmentation for simultaneous lumen and thrombus extraction is an interesting line of research. Lumen segmentation in the post-operative scenario is challenging due to contrast inhomogeneity and appearance of artifacts due to the presence of the stent-graft. Neural network based lumen segmentation approaches could solve some of the issues traditional intensity-based approaches present, and the simultaneous segmentation of the aneurysm lumen and thrombus could improve the results for both structures.  

\section*{Acknowledgements} This work has been supported by the AVICENA (ZL-2018/00966) research project, funded by the HAZITEK 2018 program from the Basque Government.

\bibliographystyle{model2-names}
\bibliography{Biblio.bib}

\end{document}